\documentclass{article}
\usepackage[english]{babel}
\usepackage[letterpaper,top=2cm,bottom=2.5cm,left=3cm,right=3cm,marginparwidth=1.75cm]{geometry}

\usepackage{amsmath}
\usepackage{multirow}
\usepackage{graphicx}
\usepackage[table,xcdraw]{xcolor}
\usepackage{caption}
\usepackage{subcaption}
\usepackage[colorlinks=true, allcolors=blue]{hyperref}
\providecommand{\keywords}[1]{
  \small    
  \textbf{\textit{Keywords---}} #1
}

\title{Enhancing Breast Cancer Diagnosis in Mammography: Evaluation and Integration of Convolutional Neural Networks and Explainable AI}
\usepackage{authblk}

\begin{document}

\author{Maryam Ahmed}
\author{Tooba Bibi}
\author{Rizwan Ahmed Khan\thanks{Corresponding author: \texttt{Rizwan.khan@shu.edu.pk}}}
\author{Sidra Nasir}
\affil{Department of Computer Science, Faculty of Information Technology, Salim Habib University, Karachi, Pakistan}

\date{} 

\maketitle
\begin{abstract}
The Deep learning (DL) models for diagnosing breast cancer from mammographic images often operate as "black boxes," making it difficult for healthcare professionals to trust and understand their decision-making processes. 
The study presents an integrated framework combining Convolutional Neural Networks (CNNs) and Explainable Artificial Intelligence (XAI) for the enhanced diagnosis of breast cancer using the CBIS-DDSM dataset. The methodology encompasses an elaborate data preprocessing pipeline and advanced data augmentation techniques to counteract dataset limitations and transfer learning using pre-trained networks such as VGG-16, Inception-V3 and ResNet was employed. A focal point of our study is the evaluation of XAI's effectiveness in interpreting model predictions, highlighted by utilising the Hausdorff measure to assess the alignment between AI-generated explanations and expert annotations quantitatively. This approach is critical for XAI in promoting trustworthiness and ethical fairness in AI-assisted diagnostics. The findings from our research illustrate the effective collaboration between CNNs and XAI in advancing diagnostic methods for breast cancer, thereby facilitating a more seamless integration of advanced AI technologies within clinical settings. By enhancing the interpretability of AI-driven decisions, this work lays the groundwork for improved collaboration between AI systems and medical practitioners, ultimately enriching patient care. Furthermore, the implications of our research extended well beyond the current methodologies. It encourages further research into how to combine multimodal data and improve AI explanations to meet the needs of clinical practice.
\end{abstract}
\section{Introduction}\label{sec1}
Breast cancer remains the predominant oncological challenge among women globally, characterized by the uncontrolled growth of abnormal cells within the mammary gland. In 2020, this malignancy was diagnosed in approximately 2.3 million women worldwide, culminating in 685,000 fatalities~\cite{iarcBreastCancer}. Computer-aided detection (CAD) systems have been developed in response to this global health issue, leveraging artificial intelligence (AI) algorithms to enhance mammogram interpretation accuracy. These systems facilitate the identification of suspicious regions within digital imaging, subsequently offering diagnostic classifications to assist physicians, thereby acting as an adjunctive opinion~\cite{moghbel2013review}.

AI methodologies employed in these systems are divided into two principal categories: conventional machine learning (ML) algorithms, which use hand-crafted features, and DL, which autonomously extract pertinent features from data~\cite{sajid2022breast}. Traditional techniques, such as the K-Nearest Neighbor and Decision Tree algorithms, exhibit limitations in managing voluminous and high-dimensional datasets due to the intricate requirement for robust feature identification and extraction, demanding extensive domain expertise~\cite{shah2022artificial}. In contrast, Deep Neural Networks (DNNs), characterized by multiple hidden layers~\cite{Bejani2021}, automatically extract features from raw inputs to yield accurate predictions, adjusting to the varied features identified. This automatic feature extraction and classification fusion attribute of DL algorithms has been instrumental in attaining unprecedented efficacy in breast cancer detection via imaging modalities~\cite{cheng2016computer}.

However, the interpretability of DNNs poses a significant challenge due to their "black-box" nature, an attribute referencing their operational opacity. These networks integrate millions of parameters across numerous layers; thus, their complex nature hinders the ability to find an explanation~\cite{arrieta2020explainable}. Although DNNs' performance in clinical diagnostics is commendable, the lack of transparency and explainability impedes their integration into routine healthcare applications~\cite{yang2022unbox}. The emergence of XAI aims to address these issues of transparency and explainability, advocating the creation of accurate AI models that are comprehensible and trustworthy to users~\cite{singh2020explainable}.

XAI is characterized as a system that generates models while maintaining high accuracy and enabling human users to comprehend and trust these models effectively. Incorporating XAI into a model boosts its credibility, aids in identifying causal relationships within the data, enhances user knowledge, empowers more assured decision-making, and fosters ethical fairness by enhancing transparency~\cite{emmert2020explainable, chereda2021explaining}. XAI techniques are fundamentally classified into two categories: transparent ML and post-hoc models. Transparent models, such as decision trees or K-Nearest Neighbors, are directly interpretable to users due to their inherent simplicity and straightforwardness~\cite{recio2020cbr}. Conversely, post-hoc models, encompassing complex DL networks like CNNs and RNNs, necessitate additional explanations to comprehend the logic behind their outputs~\cite{arrieta2020explainable}, ~\cite{kenny2021explaining}.
Explanations generated by post-hoc methods fall into two primary categories: global and local interpretability. Global interpretability provides a comprehensive view of the model’s functioning, incorporating its overall architecture, training, and data. Local interpretability, however, zeroes in on the rationale for specific predictions, analyzing relevant features and variables~\cite{singh2020explainable}. Post-hoc explanatory techniques are also further classified into model-specific and model-agnostic approaches. Model-specific methods, exemplified by Grad-CAM, are devised for particular models and are not transferable across different architectures. Model-agnostic techniques, such as LIME and SHAP, are designed for broader applicability and are capable of being deployed across a variety of models. Grad-CAM, LIME, and SHAP are among the leading explainability techniques employed in this research.

The methodology adopted in this research involves a comprehensive preprocessing of mammographic images to enhance their quality for AI analysis, followed by deploying a CNN model for classification. The selection of the CNN architecture is informed by a literature review and the model's suitability for processing high-dimensional imaging data. To address the challenge of interpretability, this study employs leading XAI techniques, including both model-specific and model-agnostic approaches, to provide insights into the decision-making process of CNN~\cite{singh2020explainable}. The efficacy of these explanatory mechanisms is assessed based on a framework that emphasizes security, transparency, and user comprehensibility~\cite{sokol2020explainability}.
This paper contributes to the growing field of AI in healthcare by providing empirical evidence on the effectiveness of combining CNNs with XAI models for breast cancer detection. It underscores the potential of such integrated approaches to not only enhance diagnostic accuracy but also foster trust and ethical fairness by making AI decisions in healthcare more transparent and understandable. ~\cite{emmert2020explainable}.

The remainder of the paper is organized as follows: Section II details the materials and methods, including data preprocessing, the CNN architecture, and the selection of XAI techniques. Section III presents the results of the classification and explanation tasks, followed by a discussion in Section IV that contextualizes the findings within the broader implications for AI in healthcare. Finally, Section V concludes the paper with a summary of the findings, limitations, and future research directions. By bridging the gap between advanced AI technologies and clinical applicability through explainability, this research aspires to pave the way for more ethical, transparent, and effective AI-assisted diagnostics in breast cancer detection, ultimately contributing to better patient outcomes and the advancement of AI in medicine.

\subsection{Explainability}\label{sec1.1}

Interpretability in ML is about comprehending how a model operates, especially when predicting variables. It provides insight into the model's prediction mechanisms without necessarily unraveling the causal relationships behind those predictions. While interpretability contributes to a model's overall explainability, not every interpretable model can be deemed fully explainable. It also encompasses the evaluation of hypothetical scenarios, allowing us to speculate on the outcomes had certain inputs been altered. The aim is to achieve a thorough understanding of the ML model, factoring in both visible and hidden elements, to construct a broad explanation of its functionality. The opaque nature of many AI models often obscures the rationale behind their outputs, complicating efforts to understand their decision-making processes. However, interpretability primarily concerns grasping the model's functional details, whereas explainability extends this by elucidating how the model reacts to new data and the implications of specific changes in its predictions. This distinction becomes particularly significant in healthcare, where the stakes of decision-making are high. A deep comprehension of how AI algorithms derive their suggestions is essential. Without such transparency, healthcare practitioners may hesitate to rely on and embrace AI technologies, fearing the inability to confirm the validity of their recommendations~\cite{10445375}. Thus, explainability and interpretability are essential for verifying the reliability and fairness of AI systems, especially in areas like healthcare, finance, and the legal system where decision-making has significant consequences.

XAI models are primarily divided into two main categories: \textbf{intrinsic} and \textbf{post-hoc}. Intrinsic models, also known as transparent models, are naturally understandable and explainable.

Conversely, post-hoc explainability techniques are further divided into model-specific and model-agnostic methods. Models lacking inherent transparency utilize post-hoc or surrogate models to clarify their decision-making processes. Model-specific explanations are custom-made to fit a model’s unique architecture, with Grad-CAM and saliency maps being notable examples. In contrast, model-agnostic explanations are more versatile and can be applied across different models to explain the prediction process, with SHAP and LIME being widely recognized methods in this category. XAI explanations can also be distinguished as local or global. Local explanations focus on understanding individual decisions made by a model, while global explanations provide insight into the overall behavior and rules of the model across all scenarios.

The subsequent sections delve into various XAI models, with an emphasis on those most commonly applied alongside ML/Deep Learning (DL) models in the context of breast cancer research. A thorough discussion of XAI models follows, including an overview of relevant research within the breast cancer domain.

\subsection{SHapley Additive exPlanations (SHAP)}\label{shap}
Introduced by Lundberg, SHAP \cite{lundberg2017} offers a coherent framework for the interpretation of model predictions, attributing significance to each feature in the context of a specific prediction. SHAP elucidates the influence of individual features on the output of any ML model, providing a tool for both global and local model understanding.

Game theory, particularly as seen through Shapley values, serves as the foundation for the SHAP methodology. It aims to distribute the prediction outcome fairly among the features, akin to dividing a payout among players based on their contribution.

\textbf{SHAP Formalism:}
The SHAP value equation is presented as follows:
\begin{equation}
\phi(f, x) = \sum_{z' \subseteq x'} \frac{|z'|!\,(M - |z'| - 1)!}{M!} \left[f_x(z') - f_x(z' \setminus i)\right]
\end{equation}

Here, $\phi(f, x)$ symbolizes the SHAP value for a model prediction $f$ with input $x$, where $z'$ denotes a subset of features, $M$ the total number of features, and $f_x(z')$ and $f_x(z' \setminus i)$ represent the model's predictions for the subset $z'$ and for the subset excluding a specific feature, respectively.

Key principles of SHAP include:

\begin{enumerate}

\item \textbf{Local Accuracy:} 
This principle asserts that the output from the explanation model $g(x')$ should align with the original model's output $f(x)$ for any given input $x$, as follows:
\begin{equation}
    f(x) = g(x') = \phi_0 + \sum_{i=1}^{M} \phi_i x'_i 
    \end{equation}
Here, $\phi_0$ is the model's base value, and $\phi_i$ are the SHAP values associated with each feature.

\item \textbf{Missingness:}
According to this principle, an absent feature (with a value of zero) in the input should not influence the model's prediction, formalized as:
\begin{equation}
x'_i = 0 \Rightarrow \phi_i = 0
\end{equation}

\item \textbf{Consistency:}
The consistency principle states that SHAP values should accurately reflect a feature's impact. If a feature's presence or constancy enhances the model's output, its SHAP value should not decrease. For any two models, $f$ and $f'$, if

\begin{equation}
f_x'(z') - f_x'(z' \setminus i) \geq f_x(z') - f_x(z' \setminus i)
\end{equation}

for all inputs $z' \in \{0,1\}^M$, then $\phi_i(f', x) \geq \phi_i(f, x)$.

\end{enumerate}

SHAP thus provides a mathematically rigorous framework for fair and consistent feature importance attribution, enhancing the interpretability of ML models.

\subsection{Gradient-weighted Class Activation Mapping (Grad-CAM)}\label{grad-cam}
Grad-CAM \cite{selvaraju2017grad}, introduced by Selvaraju et al., serves as a visual explanation method for a wide spectrum of CNN models. It operates by identifying and illuminating significant areas within an input image that are pivotal for the model's class predictions. This is achieved by calculating the gradients of any target class relative to the activations in the final convolutional layer, thereby creating a heatmap of these key regions. Distinguished from CAM, Grad-CAM boasts adaptability across various CNN architectures, enabling model interpretability without necessitating architectural modifications. As a method that discriminates between classes, Grad-CAM conducts gradient calculations and manipulates the feature maps of the concluding convolutional layer to produce meaningful visualizations.

\textbf{Grad-CAM Computation Steps:}
\begin{enumerate}
\item \textbf{Gradient Calculation:}
The first step of Grad-CAM involves the calculation of the gradient of the loss concerning the final convolutional layer's activations:
\begin{equation}
\frac{\partial L}{\partial A^k}
\end{equation}
This step discerns the contribution of each segment within the activation maps towards the model's loss, delineating their significance in the decision-making process.

\item \textbf{Global Average Pooling (GAP) of Gradients:}
Following, a Global Average Pooling (GAP) operation is applied to these gradients to ascertain the importance weights ($\alpha_k$) for each channel within the activation maps:
\begin{equation}
\alpha_k = \frac{1}{Z} \sum_i \sum_j \frac{\partial L}{\partial A^k_{i,j}}
\end{equation}
Herein, $Z = H \times W$ represents the total count of elements in the activation map $A^k$, with $\alpha_k$ elucidating the weightage of each channel's impact on the model's output.

\item \textbf{Final Grad-CAM Equation:}
The construction of the Grad-CAM heatmap is then achieved by applying these weights to the activation maps and incorporating the ReLU function:
\begin{equation}
\text{Grad-CAM}_c = \text{ReLU} \left( \sum_k \alpha_k A^k \right)
\end{equation}

\end{enumerate}
This formula emphasizes the collaborative effect of the weighted activation maps on illuminating the predictive regions of the input image. The ReLU ensures visualization of only those features positively affecting the target class, thereby aiding in the interpretability of the model’s predictive behavior.

\subsection{Local Interpretable Model-Agnostic Explanations (LIME)}\label{lime}
Ribeiro et al. introduced Local Interpretable Model-Agnostic Explanations (LIME)~\cite{ribeiro2016}, which generate explanations that are both locally accurate and comprehensible for individual predictions by simulating the model's decision boundary to be linear in the proximity of the instance being explained. It explicates an instance by fitting an interpretable model to a perturbed dataset around the input instance, shedding light on the prediction mechanisms of the original model.

\textbf {LIME Formalism:}

\begin{enumerate}
\item \textbf{Interpretable Data Representations:} 
The distinction between the actual features used by the model and those used for explanations is pivotal. For comprehensibility, explanations utilize representations that are understandable to humans, which might be different from the complex features the model employs. For text classification, this could be a binary vector denoting word presence, and for image classification, it might indicate the presence or absence of super-pixels, despite the model using complex features like word embeddings or pixel tensors.
\begin{equation}
x \in {R}^d
\end{equation}
This represents the original instance, and 
\begin{equation} 
x' \in \{0, 1\}^{d'}
\end{equation}
This is its binary vector form for interpretable representation.
  
\item \textbf{Fidelity-Interpretability Trade-off:} 
The explanation model, $g \in G$, where $G$ encompasses models such as linear models, decision trees, or lists, is chosen for its simplicity. The domain of $g$ is $\{0, 1\}^{d'}$, acting over the binary vector. The complexity of an explanation model, $\Omega(g)$, varies; for instance, it might be the depth of a decision tree or the count of non-zero weights in a linear model. The fidelity function, $L(f, g, \pi_x)$, assesses how accurately the explanation model approximates the original model $f$ near instance $x$, with $\pi_x(z)$ defining proximity to $x$. To balance interpretability and local fidelity, the optimal explanation minimizes:
\begin{equation}
\xi(x) = \arg\min_{g \in G} L(f, g, \pi_x) + \Omega(g)
\end{equation}

\item \textbf{Sampling for Local Exploration:} 
LIME approximates the loss $L(f, g, \pi_x)$, weighted by $\pi_x$, by sampling around $x'$ without presupposing any structure for $f$. This sampling constructs a dataset $Z$ of perturbed samples with labels generated by the model $f$. Optimizing equation (10) with dataset $Z$ yields a locally faithful explanation $\xi(x)$, emphasizing the need for a balance between simplicity and fidelity.

  \item \textbf{Sparse Linear Explanations:} 
  Considering $G$ as the class of linear models and $L$ as the locally weighted square loss, the aim is to derive a sparse linear model that remains interpretable. This is facilitated using a kernel function $\pi_x(z)$ to stress local proximity around $x$:
    \begin{equation}
    L(f, g, \pi_x) = \sum_{z, z' \in Z} \pi_x(z) (f(z) - g(z'))^2
    \end{equation}
To ensure interpretability, limitations on the number of features in the explanation are imposed, typically achieved through regularization methods like Lasso for feature selection, followed by least squares for weight determination, known as K-LASSO.
\begin{equation}
\Omega(g)  = \infty1[||w_g||_0 > K]
\end{equation}

\end{enumerate}
LIME strategically creates explanations that are both locally accurate and interpretable, offering insights into the operational behaviors of complex models through interpretable approximations.

\section{Related Work}\label{sec2}

Breast cancer detection often relies on ML/DL models to analyze medical imagery, such as mammograms, for diagnosis. These models, though powerful, are frequently viewed as "black boxes" due to the opaque nature of their decision-making processes. This obscurity can breed mistrust and hinder their adoption in clinical settings. Explainable AI (XAI) seeks to mitigate this problem by making the workings of these models more transparent. For instance, saliency maps are a common XAI technique that illuminates the key features within an image that influence the model's predictions. This enables healthcare practitioners to better comprehend the basis of a model's diagnostic suggestions, leading to more informed clinical judgments. Moreover, methods like SHAP (SHapley Additive exPlanations) further aid in explaining the model predictions for breast cancer detection, offering detailed explanations that healthcare professionals can rely on for making diagnosis decisions.
Moncada et al.~\cite{moncada2021explainable} benchmarked the performance of ML models such as Random Survival Forest (RSF), Survival Support Vector Machine (SSVM), and Extreme Gradient Boosting (XGBoost) against the CPH model for predicting survival outcomes in breast cancer patients. The study focused on female patients from the Netherlands diagnosed with primary invasive non-metastatic breast cancer between 2005 and 2008, highlighting the predictive prowess of XGBoost with a c-index of ~0.73, facilitated by SHAP for model explainability.

While in ~\cite{silva2023hybrid}, a hybrid model was introduced that integrates ML with explainable AI (XAI), specifically using XGBoost and SHAP for both prediction and explanation. This approach emphasized the relevance of lifestyle and biological factors, such as a high-fat diet and breastfeeding, in breast cancer risk stratification.

Moreover, the integration of radiomics and XAI has opened new avenues in the molecular subtype classification of breast cancer, with studies demonstrating the potential of imaging features to discriminate between luminal and non-luminal subtypes ~\cite{wang2022potential}. SHAP dependence plots have provided insights into the contribution of specific radiomic features to model predictions. It describes how a single feature affects the output of the LASSO SVM prediction model. The SHAP value is used to estimate the contribution of each feature to the predicted result~\cite{rodriguez2019interpretation}. 

For visual explanations of DL models' decisions in medical imaging, Grad-CAM has profound significance in breast cancer diagnosis. By generating heatmaps that highlight key regions in mammograms, Grad-CAM offers clinicians intuitive insights into the features driving model predictions. Its accessibility and ability to bridge the gap between AI-driven diagnostics and clinical practice make Grad-CAM invaluable for enhancing the trust and acceptance of AI systems in healthcare. This visual aspect of Grad-CAM significantly contributes to the interpretability and transparency of medical results, facilitating more informed clinical decisions.

Masud et al.~\cite{masud2020convolutional} employed pre-trained CNN models for classifying breast cancer from ultrasound images, where ResNet50 and VGG16 stood out for their performance, achieving an accuracy of 92.4\% and an AUC score of 0.97, respectively. The application of Grad-CAM heatmaps in their study illuminated the pivotal areas for cancer classification, thereby enhancing the model's transparency.

Subsequently, ~\cite{dong2021one} utilized DenseNet-121 for classifying primary breast lesions from 2D ultrasound images, introducing Grad-CAM visualizations as a novel tool for interpreting AI decisions. This methodology yielded impressive outcomes, with accuracy, sensitivity, and specificity rates of 88.4\%, 87.9\%, and 89.2\%, respectively, for coarse regions of interest (ROIs), and 86.1\%, 87.9\%, and 83.8\%, for fine ROIs. Such visual insights facilitated a deeper understanding of the AI’s reasoning process.

Similarly, in ~\cite{suh2020automated} authors employed DenseNet-169 and EfficientNet-B5 to automate malignancy detection, achieving accuracy levels of 88.1\% and 87.9\%, respectively. Here, Grad-CAM played a crucial role in highlighting the significant regions leading to malignancy predictions, offering insights into the importance of both the tumor and adjacent areas.

Lou et al.\cite{lou2021mgbn} introduced the multi-level global-guided branch-attention network (MGBN) for mass classification in mammograms, incorporating Grad-CAM to verify the network's accuracy in identifying mass regions. This method not only enhanced the model's reliability and interpretability but also achieved noteworthy AUC scores of 0.8375 and 0.9311 on the DDSM and INbreast databases, respectively. Building on the theme of enhancing model transparency and efficacy, Wang et al.\cite{wang2023information} took a multitask approach with their MIB-Net, which merged classification and tumor segmentation tasks. Utilizing Grad-CAM visualizations, they provided insights into how the model focuses on significant areas across various modalities, showcasing the model's capability to efficiently prioritize critical information for accurate predictions. Together, these studies highlight the evolving landscape of breast cancer diagnostics, where advanced ML models are increasingly leveraged for their precision, reliability, and interoperability.

LIME provides localized insights into the models' predictions, including breast cancer diagnosis, by highlighting specific regions of interest in mammograms contributing to the decisions. 

The authors in \cite{kaplun2021cancer} present an automated breast cancer cell image analysis system using the public BreakHis dataset\cite{spanhol2015dataset}. Utilizing Zernike image moments to extract intricate features from cancer cell images, they employ simple neural networks for binary classification (benign vs. malignant classes). Subsequently, they utilize the LIME technique for explaining the test results, highlighting significant regions responsible for the ML algorithm's decision-making process. This approach enhances the interpretability of test outcomes and justifies the algorithm's decisions based on input images, addressing the need for explainable AI in histopathology-based breast cancer detection.

Despite the advancements in utilizing XAI techniques for breast cancer detection through medical imaging analysis, a critical gap remains in the quantitative evaluation of these methods. Current literature, including seminal works by Rezazadeh et al.\cite{rezazadeh2022explainable}, Lamy et al.\cite{lamy2019hierarchical, lamy2019explainable}, and others, vividly illustrates the application of XAI to make the decision-making processes of DL models transparent. However, the absence of a unified framework for the objective measurement and comparison of XAI's effectiveness limits the ability to ascertain its actual impact on enhancing model interpretability and trustworthiness among healthcare professionals. This gap not only obscures the direct benefits of XAI in clinical settings but also hinders the strategic selection and application of XAI techniques tailored to specific diagnostic challenges in breast cancer detection. It is imperative to establish standardized metrics and benchmarks that enable the rigorous evaluation of XAI techniques across various dimensions, including their influence on diagnostic accuracy, clinician decision-making, and patient outcomes. Such a comprehensive approach would facilitate a deeper understanding of the comparative advantages of different XAI methods, guiding their optimal utilization in clinical practice.

\section{Methodology}\label{sec3}

\begin{figure}[h!]
\centering
\includegraphics[width=0.9\textwidth]{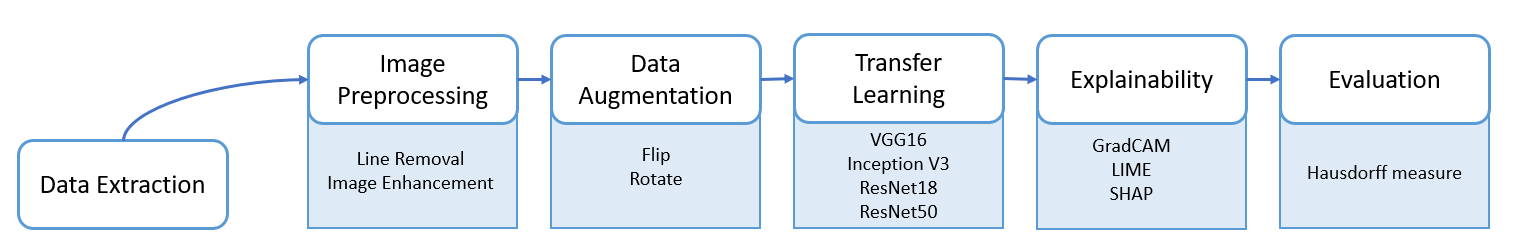}
\caption{\label{fig:syso} Breast Cancer Diagnosis system overview using CNNs and XAI.}
\end{figure}

This section outlines the comprehensive methodology adopted for the detection and classification of breast cancer using a Neural networks with XAI techniques. Given the critical nature of breast cancer as a global health issue, the focus is on developing a model that not only achieves high accuracy but also provides transparency in its decision-making process to healthcare professionals.

The methodology is structured into several key components: the dataset used for training and evaluation, the preprocessing steps to prepare the data, the adoption of transfer learning for model development, and the implementation of XAI techniques to explicate the model's output. Each component is designed to address specific challenges in the task at hand, from handling medical imaging data to enhancing the interpretability of complex DL models. By integrating these components, this methodology aims to contribute to the early detection and classification of breast cancer, providing a basis for further research and potential clinical applications, as illustrated in Figure~\ref{fig:syso}.
   

\subsection{Dataset}
The Curated Breast Imaging Subset of the Digital Database for Screening Mammography (CBIS-DDSM) is an updated and standardized version of the DDSM, containing 2,620 mammography studies categorized into malignant, benign, and normal classifications as depicted in Table~\ref{tab:model} . This extensive dataset includes 10,239 images, totaling 163.6 GB, featuring mammograms coupled with accurate pathology annotations through ROI segmentation and bounding boxes. Its comprehensive scale and verified annotations establish CBIS-DDSM as the preferred dataset for this study.

\begin{table}[ht]
\centering
\begin{tabular}{lcccc}
\hline
\textbf{Characteristic} & \textbf{Total Count} & \textbf{Categories} & \textbf{Origin} \\
\hline
Total Images & 10,239 & \multirow{2}{*}{Malignant, Benign, Normal} & \multirow{2}{*}{USA} \\
Total Subjects & 6,671 & & \\
\hline
\end{tabular}
\caption{Statistics of the Curated Breast Imaging Subset of the Digital Database for Screening Mammography (CBIS-DDSM)}
\label{tab:model}
\end{table}

In this study, 2,129 mammograms were selected from the dataset, with each corresponding mammogram's ROI extracted and converted from DICOM to PNG format. The number of ROIs varies across images, with each highlighting the critical features. All ROIs from an individual mammogram were merged into a single composite image, resulting in a dataset of 2,129 paired mammograms and ROI images. 
\begin{figure}[h!]
\centering
\includegraphics[width=0.7\textwidth]{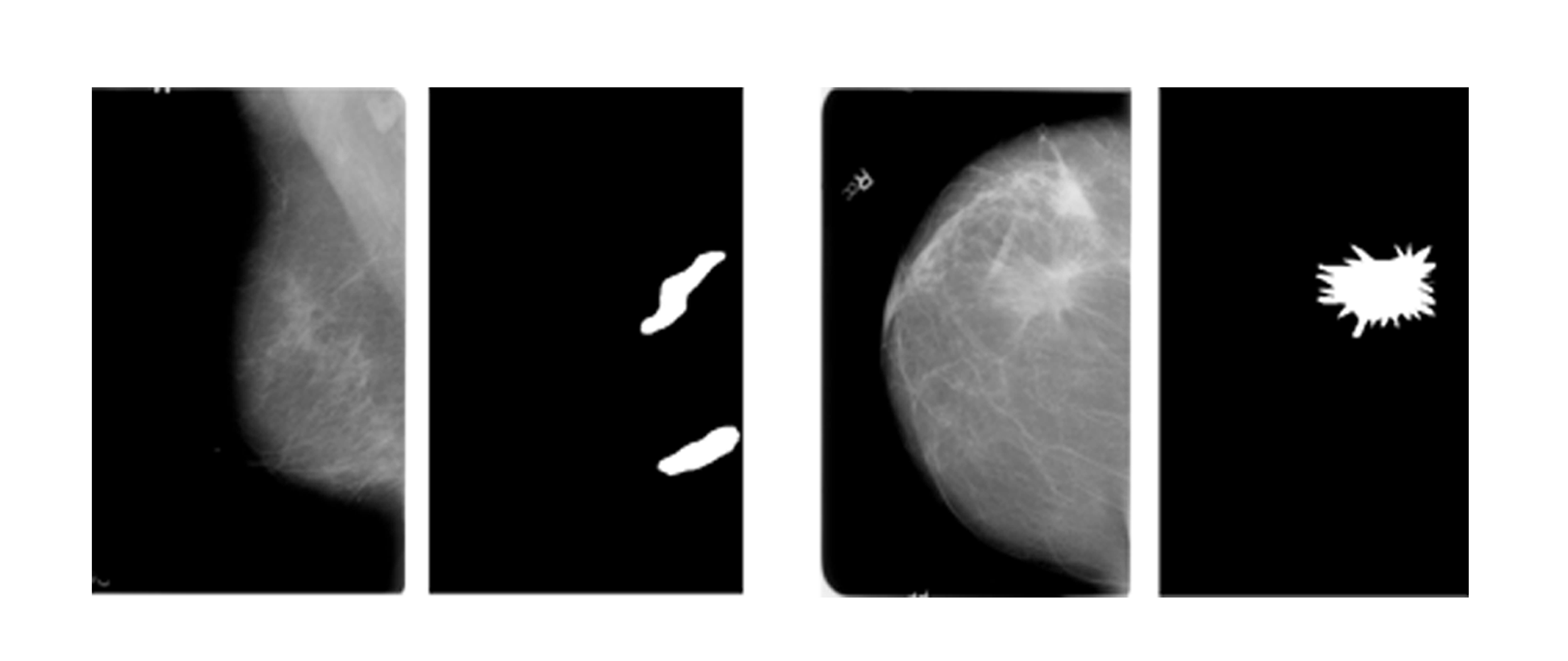}
\caption{\label{fig:benign} A benign mammogram and its Region of Interest (ROI) on the left are contrasted with a malignant mammogram and its ROI on the right~\cite{lee2017curated}.}
\end{figure}

\begin{figure}[!ht]
\centering
\includegraphics[width=0.95\textwidth]{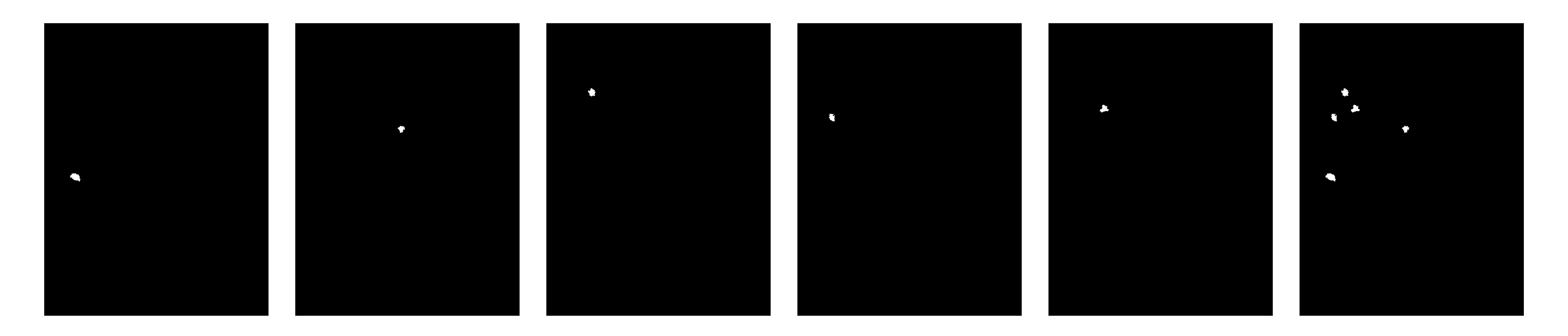}
\caption{\label{fig:combined} The five ROI images, proceeding from left to right, are merged into a single comprehensive ROI image on the right.}
\end{figure}

An illustration of this process can be seen in Figure~\ref{fig:combined}, which displays a composite of five ROIs from one mammogram. Accompanying CSV files classify these images into benign and malignant categories, comprising 1,229 benign and 900 malignant cases. Figure~\ref{fig:benign} exemplifies a benign and a malignant mammogram alongside their respective ROIs from the dataset.

\subsection{Image Pre-processing}
Preprocessing is crucial in optimizing the efficacy and computational efficiency of DL models applied to medical imaging. For this research, the mammographic images within the dataset were standardized to a resolution of 224x224 pixels. This dimensionality adjustment is followed by a rigorous preprocessing pipeline meticulously designed to expunge extraneous details while simultaneously amplifying diagnostically relevant features within the images. The preprocessing regimen encompasses three principal components: artifact reduction, line removal, and image enhancement, each underpinned by data augmentation strategies to expand the dataset beyond its original volume of 2,129 images.

\subsubsection{Artifact Reduction}
\begin{figure}[h!]
\centering
\includegraphics[width=1\textwidth]{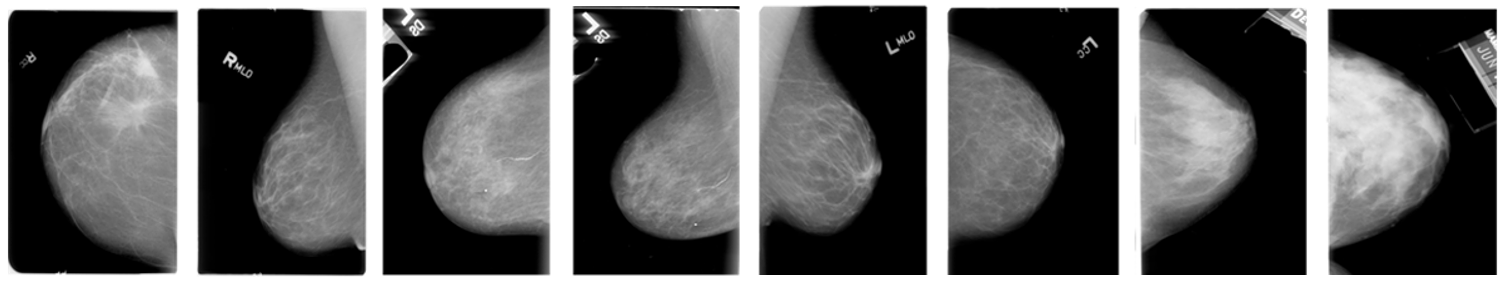}
\caption{ Mammograms displaying unwanted text (on the left) and extraneous objects (on the right).} \label{fig:textobj}
\end{figure}

\begin{figure}[h!]
\centering
\includegraphics[width=0.6\textwidth]{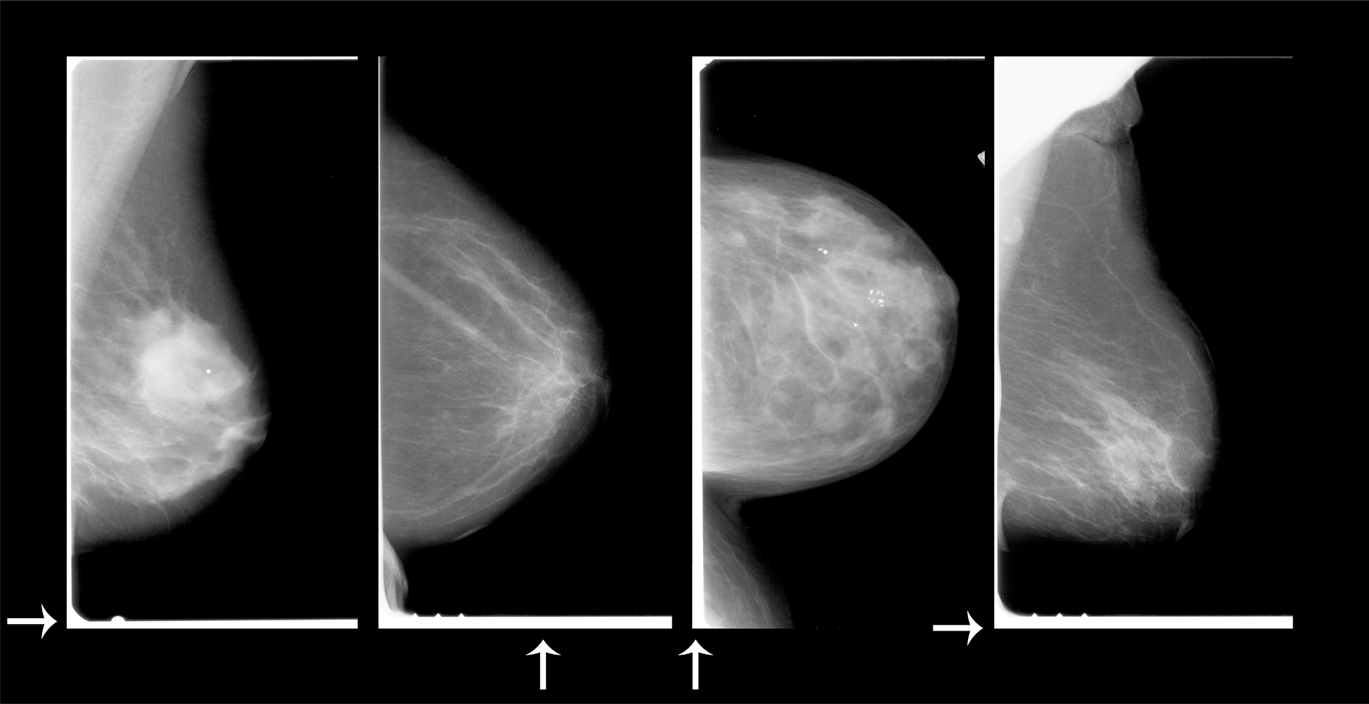}
\caption{Mammograms featuring distinct lines along their peripheral edges}\label{fig:lines}
\end{figure}

The integrity of medical images, specifically mammograms, can be compromised by the presence of non-diagnostic elements such as annotative text and extraneous objects, as depicted in Figures~\ref{fig:textobj} and \ref{fig:lines}. These artifacts, if left unaddressed, could potentially obfuscate the model's interpretative precision, leading to suboptimal detection and classification performance~\cite{lee2017curated},~\cite{ahmed2020images}. To mitigate this, artifact removal methods, which include morphological opening and contour detection,. 

\subsubsection{Line Removal}
The de-noising through binary masking proved insufficient for the complete removal of all peripheral lines, necessitating further refinement. Intensity thresholding, Gabor filtering, and morphological operations were applied to discern and eliminate residual linear artifacts. This ensures the preservation of breast tissue fidelity while removing non-essential linear distractions. ~\cite{ragab2019breast},~\cite{montaha2021breastnet18}.

\subsubsection{Image Enhancement}
The complexity of mammograms, characterized by the unclear separation between tumorous and dense tissues against a background of fatty tissues, poses a substantial challenge for interpretation based on models. This challenge is intensified in the CBIS-DDSM dataset, where dense tissues are predominantly prevalent. To counteract this, a dual-phase image enhancement strategy was adopted, incorporating Gamma correction followed by the application of Contrast Limited Adaptive Histogram Equalization (CLAHE), an evolved variant of adaptive histogram equalization~\cite{van2005morphological}. This two-step enhancement, precisely optimized for medical imaging, significantly improves the diagnostic visibility of mammograms, resulting in better model accuracy. The mathematical foundations and detailed processes of CLAHE have been comprehensively explained in~\cite{hassan2021retinex}, providing a solid foundation for its application in enhancing mammographic image quality for DL-based diagnostic assessments. 
\newline
Incorporating these preprocessing techniques within the framework has also been established by prior researchers in~\cite{lee2017curated},~\cite{ahmed2020images},~\cite{ragab2019breast},~\cite{montaha2021breastnet18},~\cite{van2005morphological},~\cite{hassan2021retinex}. This study not only adheres but also enriches the existing body of knowledge focused on refining mammographic image analysis for enhanced breast cancer detection and classification.

\subsection{Data Augmentation}

In the domain of ML applications within medical imaging analysis, data augmentation emerges as a critical preprocessing step, especially in tasks involving mammogram classification. The essence of data augmentation lies in its capacity to augment the training dataset through a series of transformations, such as flipping, rotating, and adjusting image properties. This process introduces a broader range of scenarios for the model to learn from, significantly enhancing its robustness and preventing overfitting. It is vital in medical imaging, where the ability to accurately recognize diverse patterns and anomalies can dramatically affect diagnostic outcomes. The goal of implementing specific augmentation strategies is to enrich the dataset with complexities and variabilities reflective of real-world conditions, thereby improving the model's diagnostic performance.

Before the augmentation process begins, a crucial step involves dividing the collection of mammographic images into training and testing groups, adhering to a 90:10 split. This division is essential for ensuring a balanced representation of both benign and malignant cases in the training and testing phases and maintaining the integrity of the dataset. A balanced representation aids in preventing model bias, ensuring that the diagnostic capabilities of the ML model are developed uniformly across different types of cases.

Upon establishing a structured and balanced dataset, the focus shifts to enhancing the robustness and generalizability of the model. This enhancement is achieved by applying a series of data augmentation techniques to the training set. The deployment of these techniques aims to increase the size of the training dataset by generating modified versions of the images, thereby reducing overfitting and improving the model’s ability to generalize from training to unseen data.

\begin{figure}[!htb]
\centering
\includegraphics[width=0.9\textwidth]{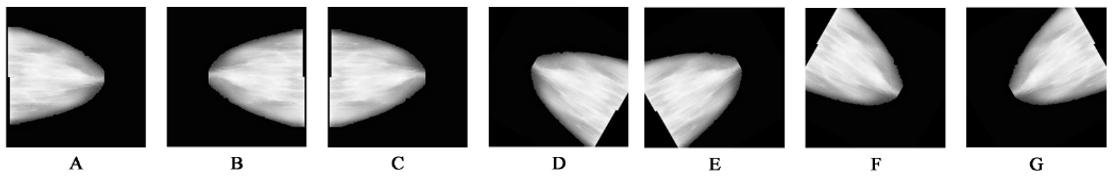}
\caption{\label{fig:augmented} Augmentation techniques used (A: vertical, B: horizontal, C: horizontal vertical, D: rotate -30° horizontal, E: rotate -30°, F: rotate 30°, G: rotate 30° horizontal).}
\end{figure}
To address this, seven distinct data augmentations were implemented on the training dataset. which include horizontal flipping, vertical flipping, and combined flipping in both directions and rotations (both positive and negative 30 degrees), with and without subsequent flipping as shown in Figure~\ref{fig:augmented}. Such comprehensive augmentation ensures the model is exposed to a wide array of image orientations and variations, mirroring the diversity encountered in real-world diagnostic settings.

\subsection{Transfer Learning}
The approach involves the use of models pre-trained on large datasets, which are then slightly adjusted or fine-tuned to perform specific tasks, such as identifying abnormalities in mammogram images. Transfer learning is particularly beneficial in situations where annotated medical datasets are scarce. By leveraging knowledge acquired from related tasks, it allows for the efficient training of robust models on smaller datasets, simply by modifying the model's final layers to adapt to the new task. This technique's capacity to overcome the challenges posed by limited data availability has been well documented in the literature of~\cite{Khan2019}. The ability to transfer learned patterns from one domain to another significantly enhances model performance and accelerates the development process in data-constrained environments.
Convolutional neural Networks (CNNs) are at the forefront of image analysis technologies, with several architectures standing out for their performance. Notable examples include VGG \cite{simonyan2014very}, Res-Net \cite{he2015deep}, Dense-Net \cite{huang2017densely}, Inception \cite{szegedy2015going}, and Alex-Net \cite{krizhevsky2017imagenet}. Each architecture offers distinct advantages, making them suitable for a wide range of applications in deciphering image data.
This study aimed to evaluate the performance of various transfer learning models on a dataset curated for mammographic image analysis. The models tested were VGG16, Inception V3, Res-Net18, and Res-Net50. Consistency in dataset usage and data split ratios ensured the reliability of performance comparisons. The objective was to identify the model that The evaluations revealed variations across the tested models, with Res-Net50 emerging as the most effective, achieving a test accuracy of 72\%. Enhancement of this model through fine-tuning—a process of re-adjusting pre-trained weights for specific tasks as detailed in~\cite{tajbakhsh2016convolutional}—led to improved accuracy of 76\% as displayed in Table~\ref{tab:model_performance}.

\begin{table}
\centering
\begin{tabular}{lcccc}
\hline
\textbf{Model} & \textbf{Epochs} & \textbf{Batch Size} & \textbf{Train Accuracy} & \textbf{Test Accuracy} \\
\hline
VGG16 & 50 & 32 & 0.92 & 0.56 \\
Inception V3 & 20 & 32 & 0.67 & 0.56 \\
Res-Net18 & 10 & 32 & 0.71 & 0.58 \\
Res-Net50 & 50 & 32 & 0.95 & 0.72 \\
Res-Net50 (Fine Tuned) & 100 & 32 & 0.95 & 0.76 \\
\hline
\end{tabular}
\caption{Performance comparison of different CNN models on the CBIS-DDSM dataset.}
\label{tab:model_performance}
\end{table}

The fine-tuned Res-Net50 model is instrumental in the integration of XAI within the mammographic analysis. XAI aims to increase transparency in AI decision-making processes, making AI-driven diagnoses more understandable. This is particularly important in the medical sector, where clear diagnostic reasoning can significantly affect patient outcomes.
The research findings demonstrate the significant impact of transfer learning on improving the accuracy of AI models for medical image analysis. Additionally, this study lays the foundation for advancements in Explainable AI (XAI) to bridge the gap between AI's capabilities and human interpretability. This increased understanding has the potential to encourage better collaboration between healthcare professionals and AI systems, ultimately leading to improved patient care.

\section{Results}\label{sec4}

In the exploration of advanced diagnostic methodologies for breast cancer, the integration of AI within mammographic analysis has shown promising potential to augment the precision and efficiency of early detection. This section delves into the empirical findings derived from the application of various interpretability techniques— Grad-CAM, LIME, and SHAP values—on mammographic images. These techniques offer a window into the AI model's cognitive process, revealing how it discerns between benign and malignant findings. Through a meticulous examination of heatmaps, LIME masks, and SHAP value plots, our study explains the model's ability to identify and prioritize areas of concern within mammographic images. This not only enhances our understanding of AI-driven diagnostics but also paves the way for improvements in the interpretability and reliability of AI models in medical imaging.
\\
\begin{figure}[!htb]
\centering
\includegraphics[width=0.6\textwidth]{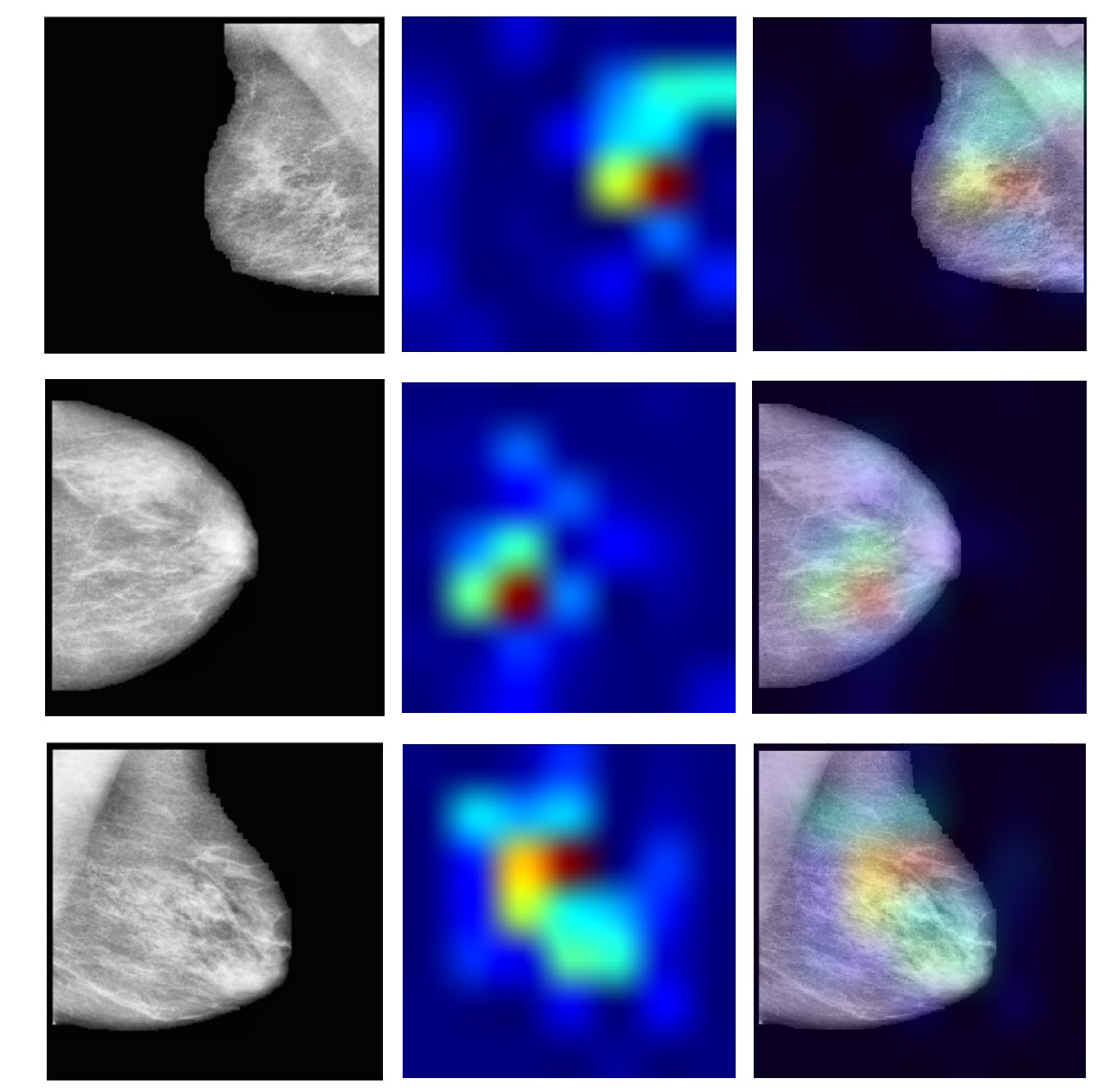}
\caption{\label{fig:heatmaps} Grad-CAM results (Left column: Original image, Center column: heatmap generated by Grad-CAM, Right column: heatmap superimposed on image)}
\end{figure}

Upon analyzing the mammographic images with heatmaps (Grad-CAM), it became evident that the model assigns heightened importance to certain regions, visibly marked by an intensified color spectrum, where red hues denote areas of critical diagnostic value. Notably, these regions often align with denser tissue or anomalies, suggesting that the model's algorithm is proficient in detecting subtle yet clinically significant patterns that might escape the unaided eye. For instance, in Figure~\ref{fig:heatmaps}, the brighter areas correlate with the precise locations of tissue irregularities, underscoring the model's potential in highlighting zones warranting closer examination for the presence of malignancies.
\\
\begin{figure}[h!]
\centering
\includegraphics[width=0.6\textwidth]{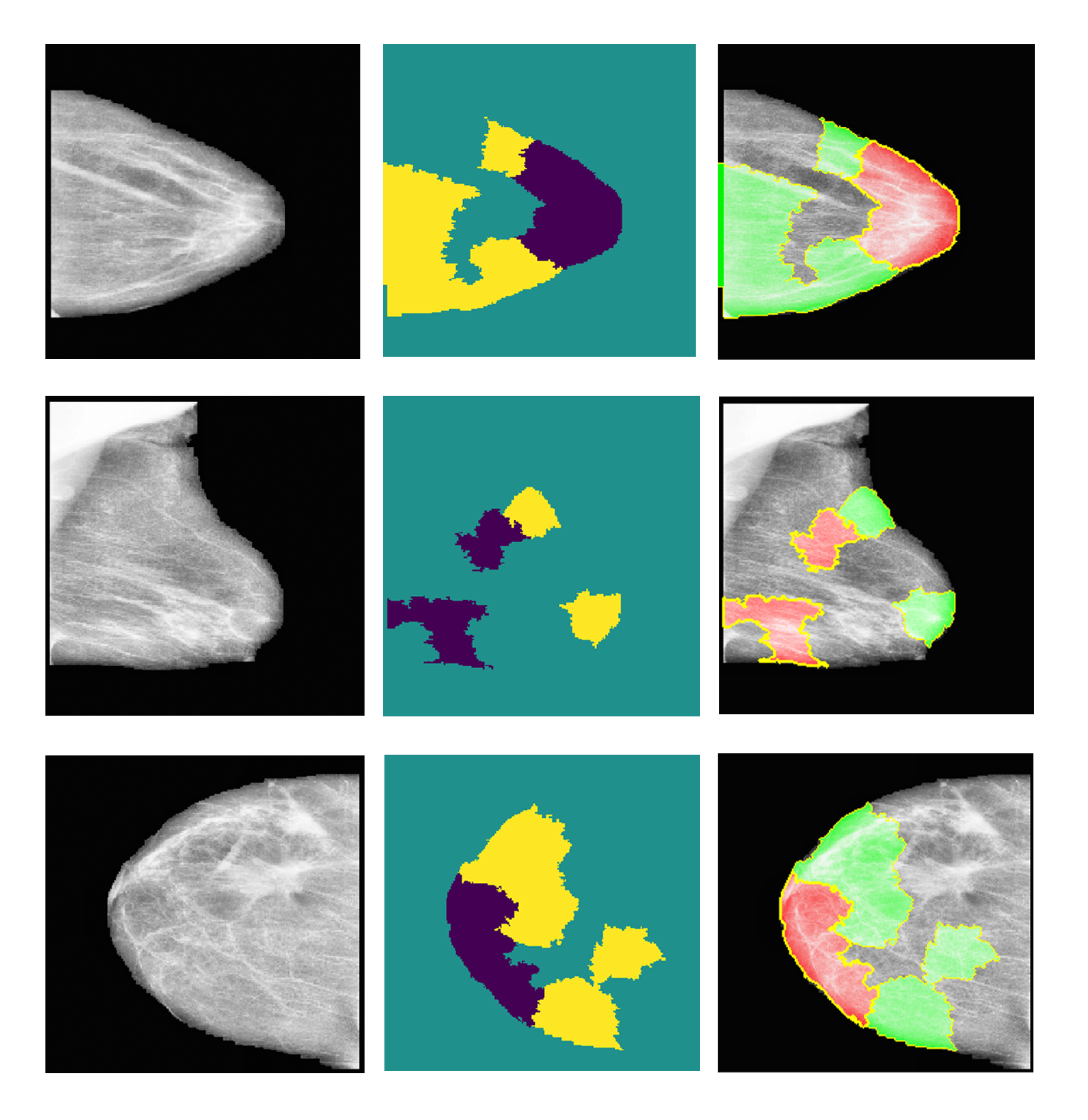}
\caption{ LIME results (Left column: Original image, Center column: segmentation result generated by LIME, Right column: important segments superimposed on image)}
\label{fig:limeresults}
\end{figure}
\\
The application of LIME masks to the original mammograms further refines our understanding of the model's diagnostic process. By assigning distinct colors to different segments, the LIME analysis dissects the mammogram into regions of varying relevance to the model's classification decision. This segmentation reveals that the model not only considers the overall tissue density but also pays detailed attention to specific areas that exhibit unusual textural or structural characteristics. For example, Figure~\ref{fig:limeresults} shows how segments marked by LIME correspond to areas where a radiologist might detect signs of pathological changes, thereby validating the model's accuracy in identifying key diagnostic features.
\begin{figure}[h!]
\centering
\includegraphics[width=0.55\textwidth]{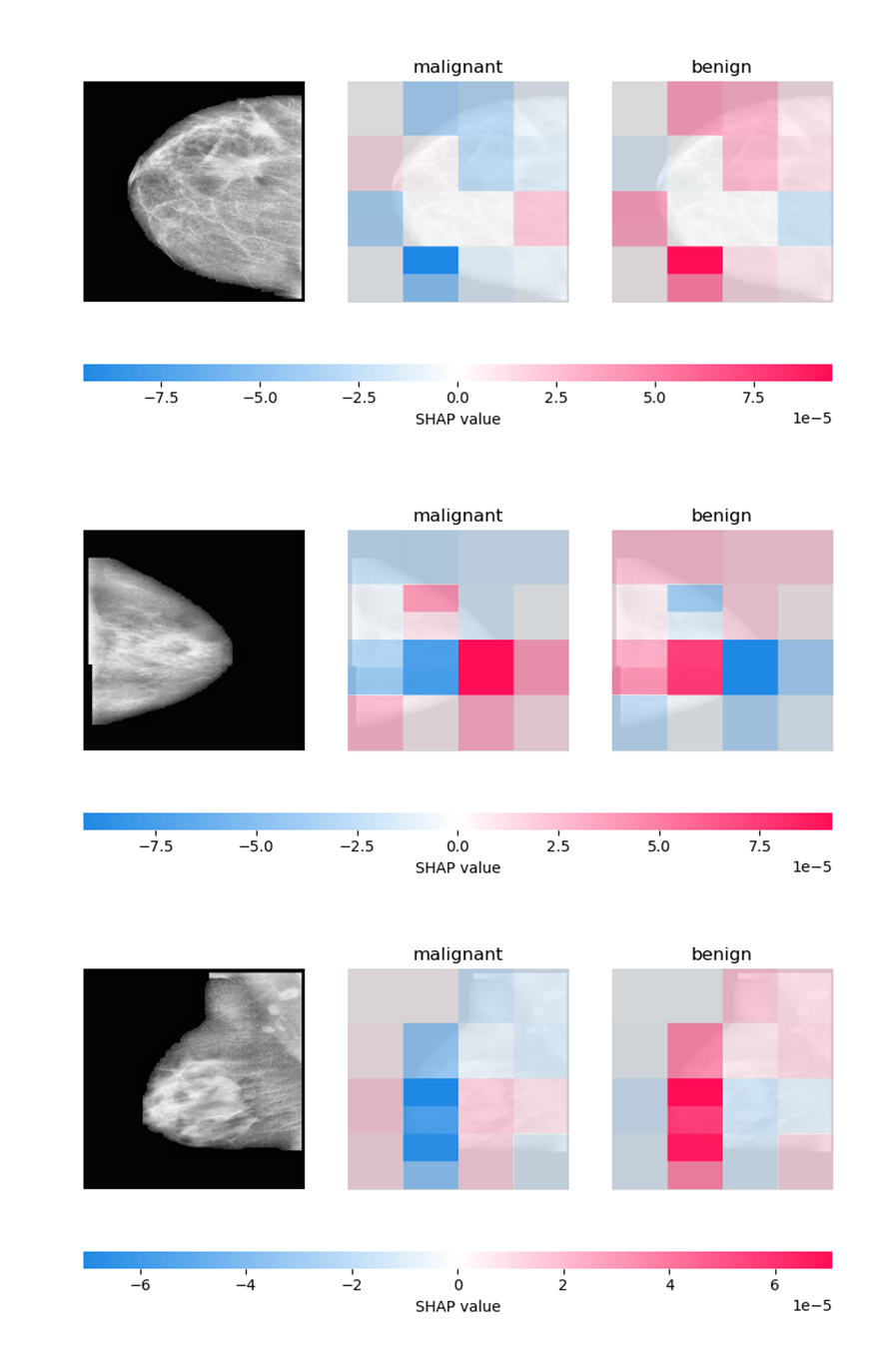}
\caption{\label{fig:SHAP}Results of SHAP}
\end{figure}
SHAP value plots offer a granular view of the model's reasoning by displaying the influence of individual features (pixels or regions) on the classification outcome. Through a color-coded representation—blue for features pushing the classification towards benign and red for those indicating malignancy—these plots articulate the model's internal assessment of each region's contribution to the final diagnosis. In the context of breast cancer detection, areas with high red SHAP values not only confirmed the presence of malignancy but also aligned with clinically relevant markers of cancerous tissue. Figure~\ref{fig:SHAP} illustrates this phenomenon, where the predictive significance of each pixel is quantified, offering a direct link between the model's interpretation and identifiable pathological features.

\section{Evaluation}\label{sec5}
Explainability methods such as Grad-CAM, LIME, and SHAP are vital for deciphering the workings of DL models. However, assessing the safety and reliability of these explanations is equally crucial. Our evaluation, based on the explainability factsheet by Sokol et al. \cite{sokol2020explainability}, primarily focuses on the fourth dimension: safety. This dimension encompasses four critical factors: {\em information leakage}, {\em explanation misuse}, {\em explanation invariance}, and {\em explanation quality}. We have systematically evaluated all four of these factors.

\subsection{Information Leakage}
To ensure the safety and security of any AI system, it is important to consider how much knowledge about the model itself is revealed by the information provided in the explanation. An example of information leakage can be the explanations of a k-nearest neighbor model, which can reveal the information of the training data points of the model. However, when this factor is considered for our model, it is not a concern as we use DL models, which are essentially black-box models and require explanations to understand the results. The explanations provided by our chosen methods of Grad-CAM, LIME, and SHAP also do not reveal any information about the architecture of the model itself; instead, they help us understand the regions of interest chosen by the model in each sample image. Each output of the models provides visual information that highlights the most important regions in the image. 

\subsection{Explanation Misuse}

The factor of information misuse comes into consideration based on the information in the model that is leaked via explanations. A model can be stolen or replicated if considerable information about it can be gathered through the explanations. However, this factor does not apply to our model, as our methods of explanation do not leak any information regarding the architecture of the model. \\

\subsection{Explanation Invarience}

The third factor in the dimension of safety is explanation invariance. The primary objective of explanations is to develop an understanding of humans for each sample. Hence, an explainable system is expected to be stable and consistent. \\

\begin{figure}[h!]
\centering
\includegraphics[width=0.95\textwidth]{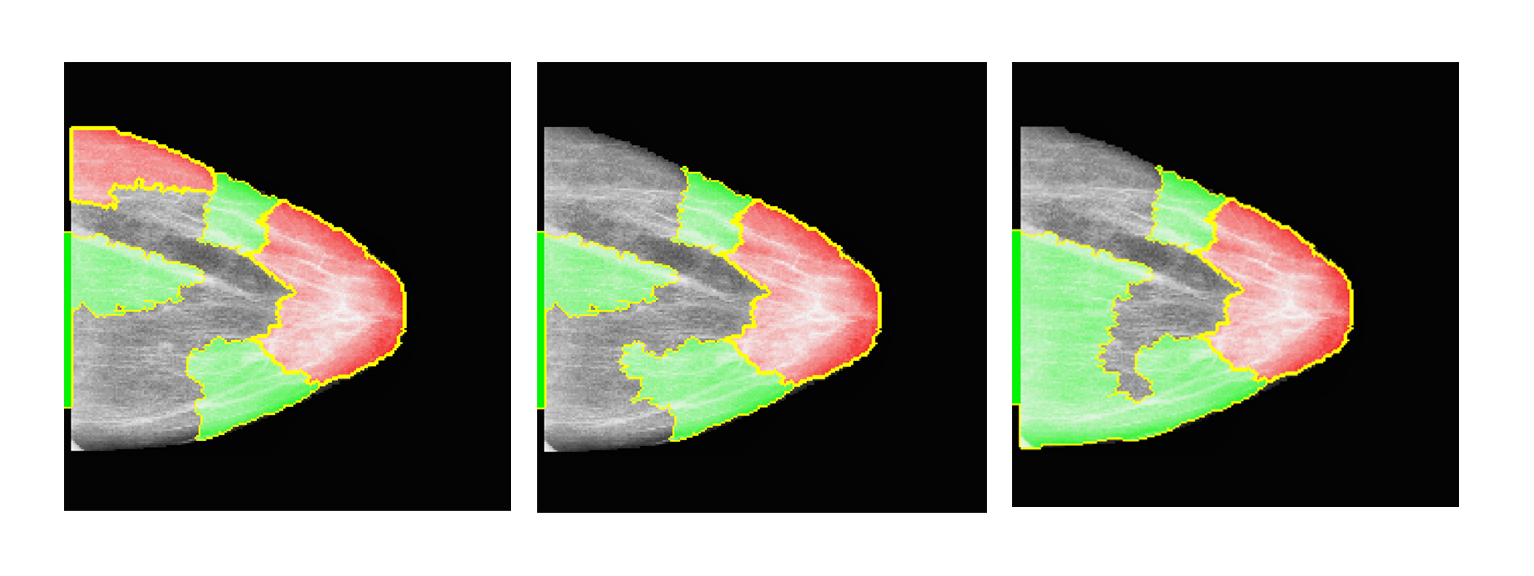}
\caption{\label{fig:samelime} Multiple LIME results for the same mammogram}
\end{figure}

The factor of {\em stability} defines that explanations of the same prediction by a model generated multiple times should result in the same explanation as well. This was evaluated in our model as well by implementing the three XAI techniques multiple times on the same image. An explanation generated by LIME results from the process of random perturbation, which generates simulated data in the surroundings of an instance for a simple linear classifier to predict and results in instability as it leads to the generation of different explanations for the same prediction \cite{zafar2019dlime}. Hence, while Grad-CAM and SHAP are stable and produce the same explanation for each prediction, the explanation provided by LIME is unstable, which has been shown in Figure~\ref{fig:samelime}. 

\begin{figure}[h!]
\centering
\includegraphics[width=0.5\textwidth]{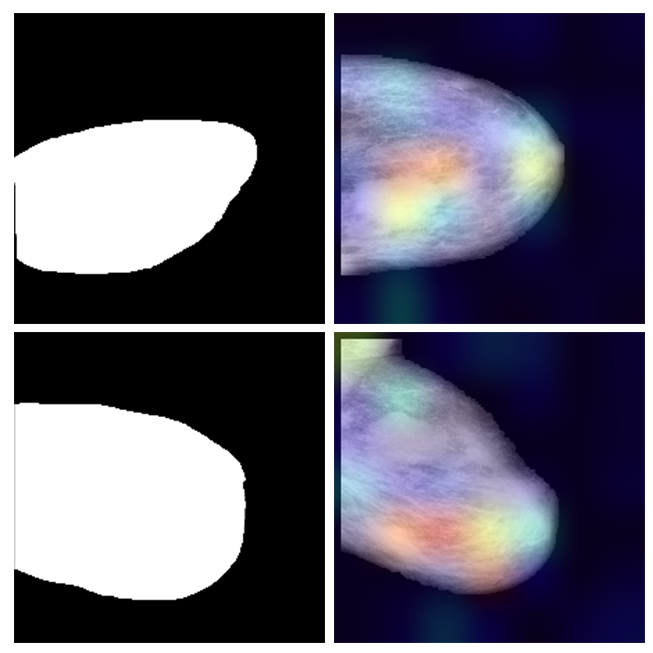}
\caption{\label{fig:similargradcam} Consistency of Grad-CAM results (left column: ROI images, right column: Grad-CAM results)}
\end{figure}

The factor of {\em consistency} defines that explanations of a fixed model for similar data points should also be similar. We have evaluated the consistency of our model by using the ROI data provided in the original CBIS DDSM dataset and checking whether images with similar regions of interest have similar XAI results. The Figure~\ref{fig:similargradcam} shows two similar ROI images with similar Grad-CAM results, which shows the consistency of Grad-CAM. SHAP works by assigning each feature an importance value for a particular prediction \cite{lundberg2017unified}. Hence, similar datapoints are assigned similar importance values, which shows the consistency of SHAP. The instability of LIME also makes LIME inconsistent as it generates different explanations for the same prediction; hence, comparison of similar data points does not produce similar explanations. \\

\subsection{Explanation Quality}

\begin{figure}[h!]
\centering
\includegraphics[width=0.8\textwidth]{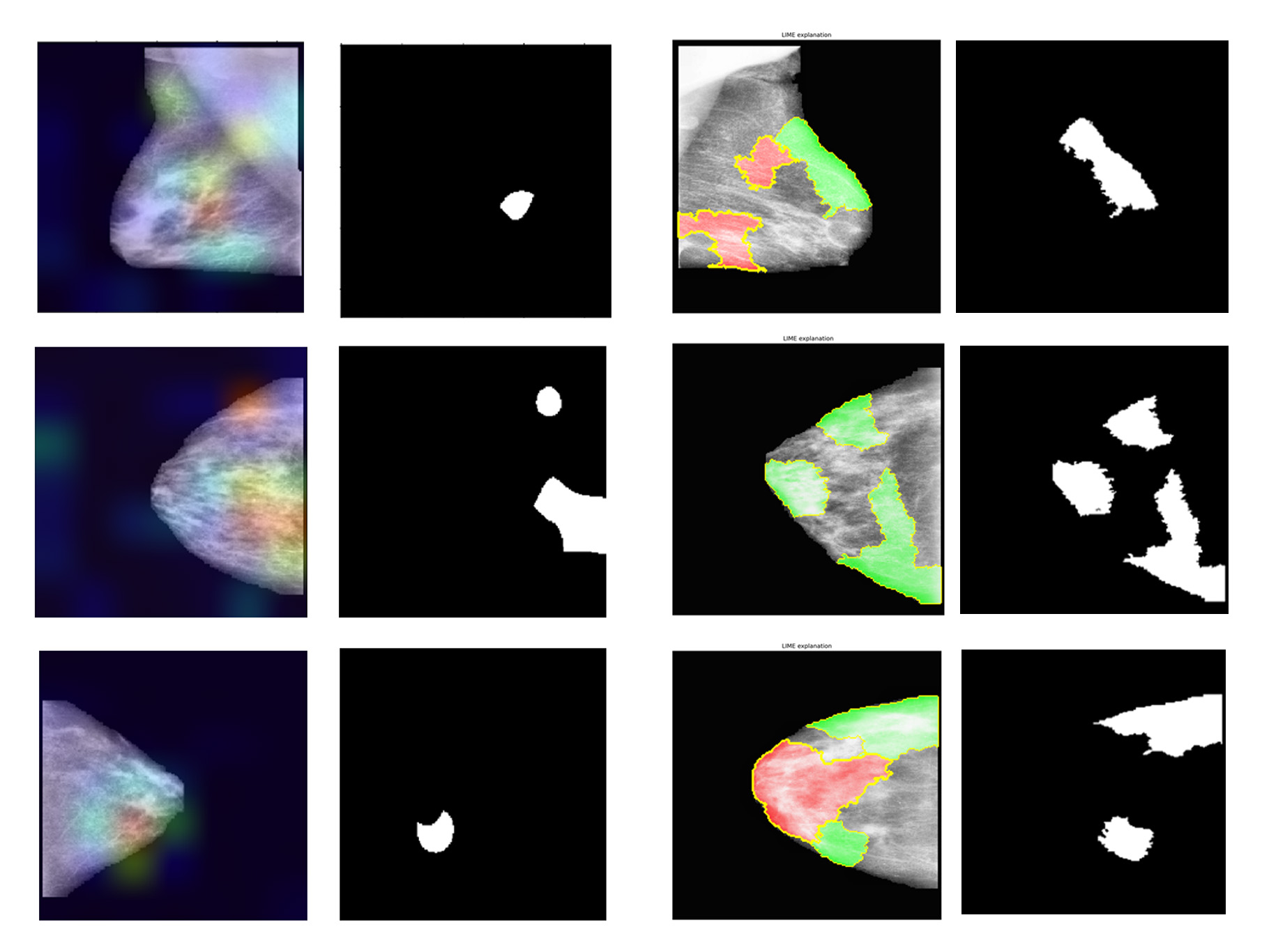}
\caption{\label{fig:masks} Grad-CAM (left) and LIME (right) results with binary Mask}
\end{figure}

The final safety measure is the explanation quality, which emphasizes the correctness of the explanation provided. While we have achieved the explainability of the model, it is necessary to analyze the correctness of the explanations provided by the different XAI functions to understand the performance of the black-box DL model. To analyze the explanations provided by XAI techniques for the performance of our model, the ROI images of the CBIS DDSM dataset have been used to compare the areas of importance highlighted by verified pathologists with the areas highlighted by XAI techniques. The ratio of similarity between the results of XAI and the original ROI images indicated how well the model is detecting the correct parts of the images for its classification. To create this ratio, we have created binary masks of the explanations provided by Grad-CAM and LIME which are shown in Figure~\ref{fig:masks}. 

\begin{figure}[h!]
\centering
\includegraphics[width=0.99\textwidth]{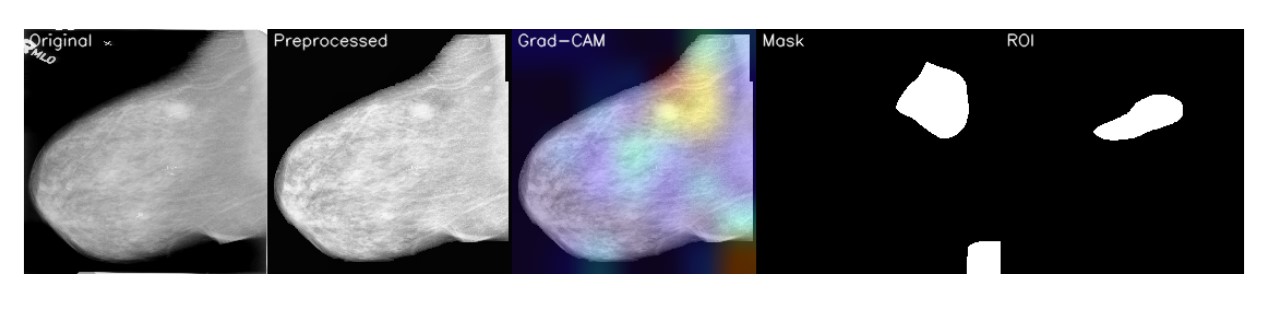}
\caption{\label{fig:mask} Grad-CAM results with Mask and ROI}
\end{figure}

Figure~\ref{fig:mask} shows the original mammogram, the pre-processed image, the Grad-CAM output, the mask generated from the Grad-CAM heatmap, and the original ROI from the dataset. \\

\begin{figure}[h!]
\centering
\includegraphics[width=0.95\textwidth]{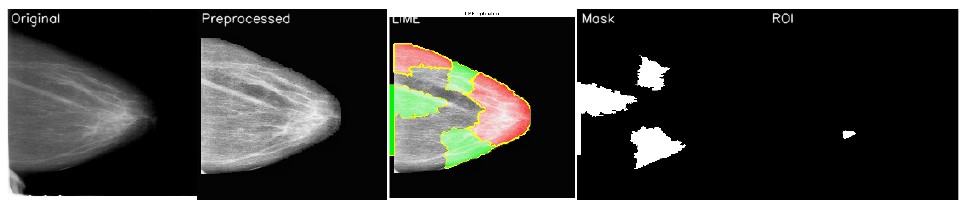}
\caption{\label{fig:limeconcat} LIME results with Mask and ROI}
\end{figure}
Figure~\ref{fig:limeconcat} shows the original mammogram, the pre-processed image, the LIME output, the mask generated from the LIME segments, and the original ROI from the dataset.

The binary masks are then compared to the ROI images using Hausdorff distance. Hausdroff distance is a common measure for comparing the dissimilarity between image segmentation and point sets \cite{taha2015efficient}. This method is commonly used in the medical domain to evaluate segmentation in medical images such as computed tomography (CT) and magnetic resonance imaging (MRI) \cite{besl1992method}. The Hausdorff distance $\check{H}$ between two point sets $A$ and $B$ is the maximum distance between each point, as shown in the following equation \cite{taha2015efficient}. 
\begin{equation} \check{H}(A,B) = max_{x \in A}\lbrace min_{y \in B}\lbrace ||x,y||\rbrace \rbrace, \end{equation}
The degree of resemblance is greater when the value of the Hausdroff distance between two shapes is smaller \cite{kim2013computing}. The range of Hausdorff distance values depends on the size and complexity of the images being compared. In general, the Hausdorff distance values range from 0 to infinity, where a value of 0 indicates that the two binary masks are identical, and larger values indicate increasing differences between the two masks. After the Hausdroff distance for each image is calculated, the mean distance is calculated to analyze the performance of the model. A lower mean Hausdorff distance indicates a higher level of similarity between all the images, while a higher mean Hausdorff distance indicates a greater level of dissimilarity. \\
When comparing the ROI images with the masks produced by Grad-CAM, the Hausdorff distance values of our dataset range from 1 to 133, with the mean Hausdorff value being 18.
When comparing the ROI images with the masks produced by Grad-CAM, the Hausdorff distance values of our dataset range from 1 to 160, with the mean Hausdorff value being 86. This indicates that the performance of the Grad-CAM explainability technique is better than the explainability provided by LIME.

\section{Conclusion}\label{sec6}
The study has explored the symbiosis of CNNs and XAI within the context of mammography for breast cancer diagnosis. The core of our research did not only improved the predictive accuracy of CNNs through the deployment of a fine-tuned ResNet50 model but also highlighted the blackbox, a void often criticized in DL methodologies, by integrating XAI techniques—namely Grad-CAM, LIME, and SHAP. These methods addressed the "black box" nature of DL, offering insightful visual and quantitative interpretations of the model's decision-making processes.

Our findings highlight the potential of combining CNNs with XAI to not only advance the accuracy of breast cancer diagnosis but also to bridge the gap between AI technologies and clinical applicability. By making AI decisions more understandable, we pave the way for enhanced collaboration between AI systems and medical professionals, ultimately contributing to more informed and trustworthy patient care.

Despite the promising results, this study acknowledges the limitations inherent in our approach, including the variability in explanation consistency observed with certain XAI techniques, notably LIME. Such limitations highlight the necessity for continuous refinement of both models and explanation methodologies to ensure that they meet the rigorous demands of clinical application.

In the future, we have envisioned several trajectories for future research. Firstly, exploring more DL architectures and training techniques could further enhance diagnostic accuracy. Secondly, developing more robust and consistent XAI methodologies will be crucial to improving the reliability of AI explanations. Thirdly, collaboration with medical professionals in designing AI tools can ensure that these technologies align with clinical needs and practices. Finally, extending our research to include multi-modal data, such as patient history and genetic information, could offer a more holistic approach to breast cancer diagnosis.

\bibliographystyle{ieeetr}


\end{document}